\documentclass[runningheads]{llncs}
\usepackage{caption}
\usepackage{graphicx}
\usepackage{pgf-pie}  
\usepackage{algpseudocode}
\usepackage{xcolor}
\usepackage{mathtools}
\usepackage{subcaption}
\usepackage{soul} 
\newcommand{\comment}[1]{}
\usepackage{tabularx}
\usepackage{array,ragged2e}
\newcolumntype{P}[1]{>{\RaggedRight\arraybackslash}p{#1}}

\newcommand{\COMMENT}[1]{$\triangleright$  #1}

\DeclarePairedDelimiter\floor{\lfloor}{\rfloor}

\begin{document}
\title{Fast Hand Detection in Collaborative Learning Environments\thanks{ This material is based upon work
    supported by the National Science Foundation under the AOLME
    project (Grant No.  1613637), the AOLME Video Analysis project
    (Grant No.  1842220), and the ESTRELLA project (Grant
    No. 1949230).  Any opinions or findings of this paper reflect the views of the authors. They do not necessarily reflect the views of
    NSF.  }}
%
%
\author{
  Sravani Teeparthi\inst{1} \and
  Venkatesh Jatla\inst{1} \and
  Marios S. Pattichis\inst{1} \and
  Sylvia Celed\'{o}n-Pattichis\inst{2} \and
  Carlos L\'{o}pezLeiva\inst{2}
}
\authorrunning{Teeparthi et al.}
%
\institute{
  The University of New Mexico, Albuquerque, NM, USA \\
  \email{\{steeparthi, venkatesh369, pattichi\}@unm.edu}\\
  \and
  Department of Language, Literacy, and Sociocultural Studies\\
  \email{\{sceledon, callopez\}@unm.edu}
}
\maketitle              

\begin{abstract}
 Long-term object detection requires the integration
    of frame-based results over several seconds.
 For non-deformable objects, long-term detection
    is often addressed using object detection followed
    by video tracking.
 Unfortunately, tracking is inapplicable to objects that undergo dramatic changes in appearance from frame to frame.
 As a related example, we study hand detection over long video recordings in collaborative learning environments.
 More specifically, we develop long-term hand detection methods that can deal with partial occlusions and dramatic changes in appearance.
    
 Our approach integrates object-detection, followed by time projections,
    clustering, and small region removal to provide
    effective hand detection over long videos.
 The hand detector achieved average precision (AP) of $72\%$ at 0.5
    intersection over union (IoU). 
 The detection results were improved to
    $81\%$ by using our optimized approach for data augmentation. 
 The method runs at $4.7\times$the real-time with AP of $81\%$ at 0.5 intersection over the union.
 Our method reduced the number of false-positive hand detections by 80$\%$ by improving IoU ratios from 0.2 to 0.5. The overall hand detection system runs at $4\times$ real-time.

\end{abstract}
\begin{figure}[t]
    \centering
    \subfloat[Sample video frame showing
    fully visible hands, occluded hands, and
    hands belonging to other groups.]
    {
        \label{subfig:problem_dramatic_change1}
    \includegraphics[width=0.45\textwidth]{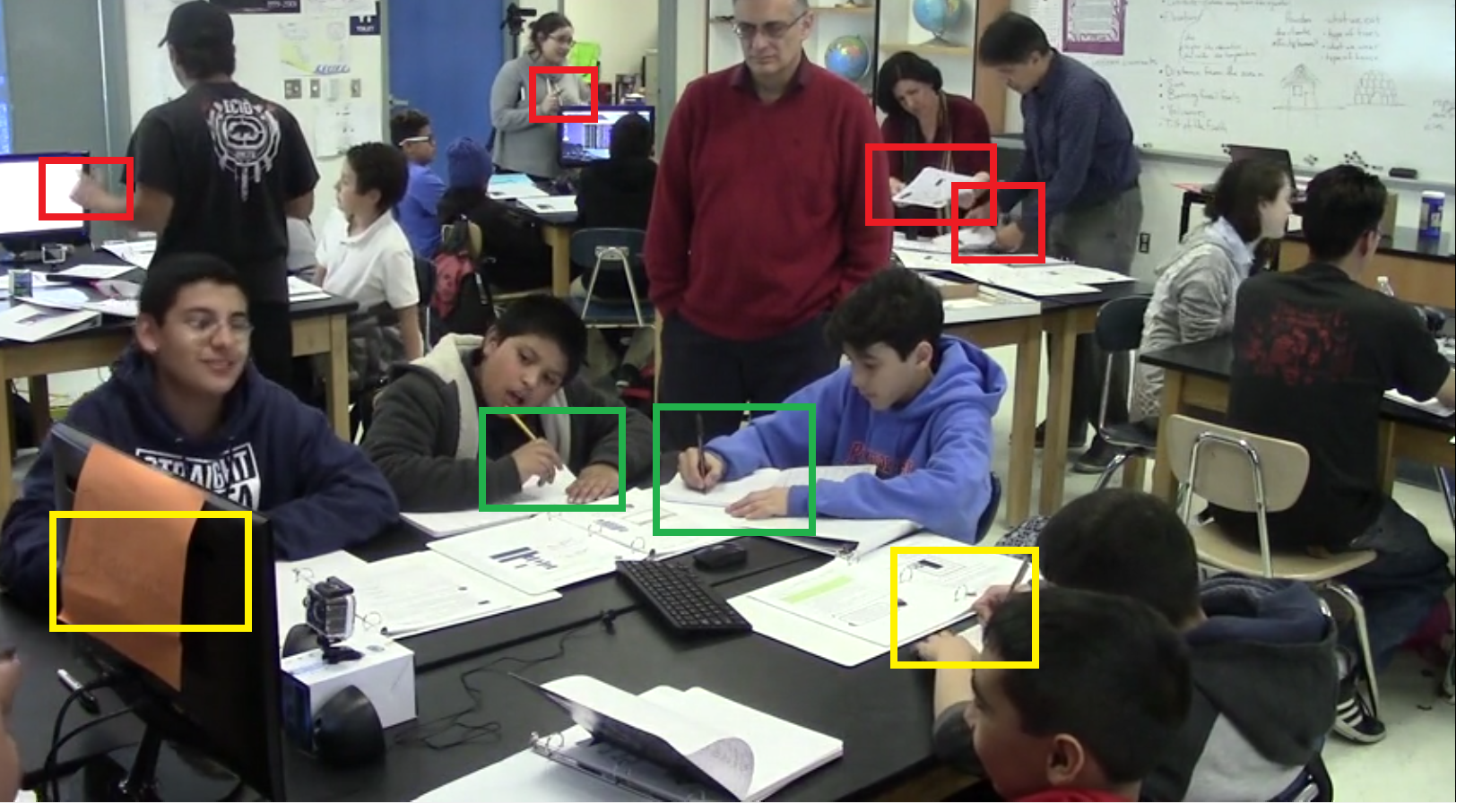}
    }\hspace{2mm}
    \subfloat[Sample video frame occuring 2 seconds
    after the frame in (a). On the lower-left, a new
    set of hands appears.]
    {   
    \label{subfig:problem_dramatic_change2}
    \includegraphics[width=0.45\textwidth]{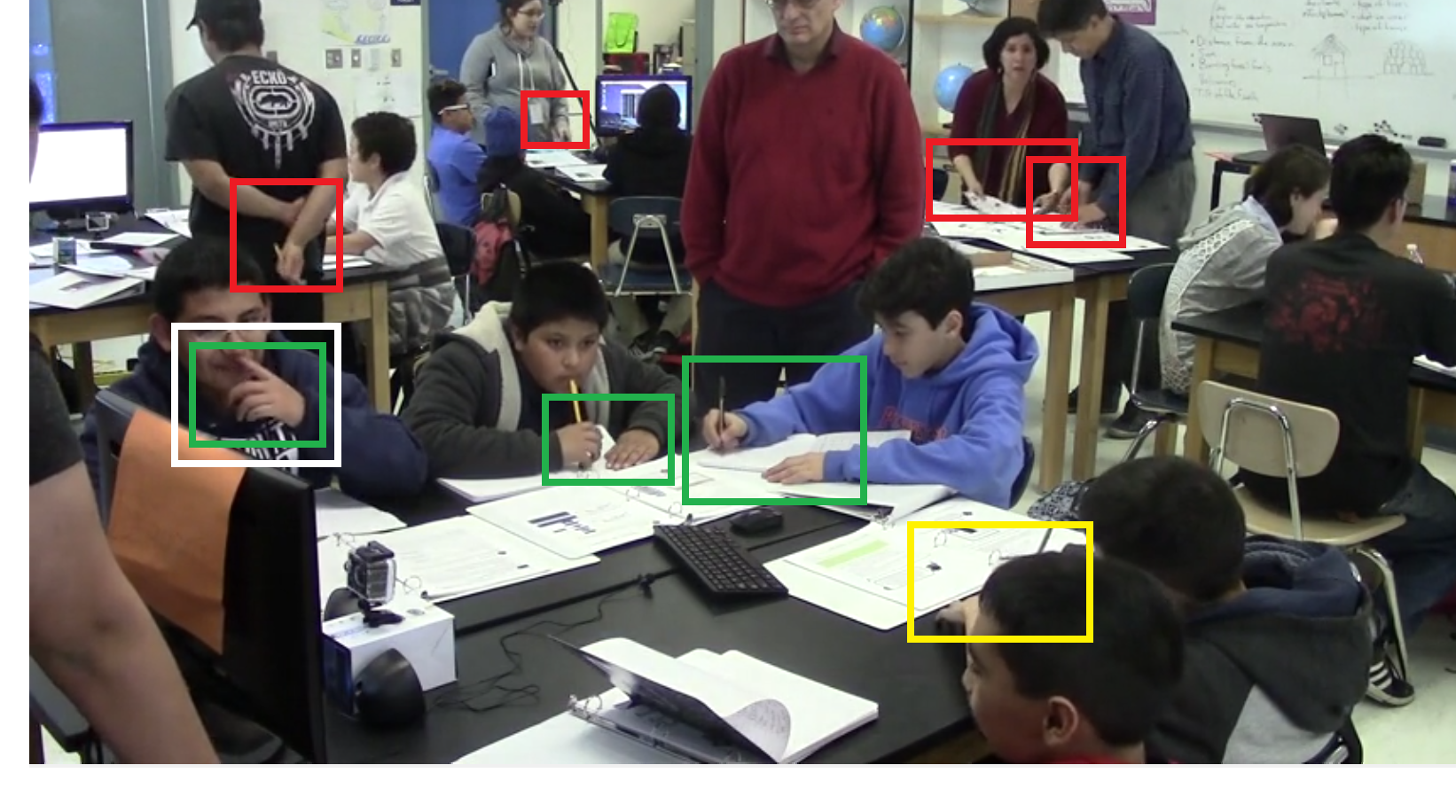}
    }
    \caption{Hand detection in collaborative learning environments.
    The problem is restricted to detecting student hands
        that are nearer to the camera.
    We use green bounding boxes to identify unobstructed hands that
        need to be detected.
    We use yellow bounding boxes to identify occluded hands
       that need to be detected through projection methods.
    We use red bounding boxes to identify hands that belong
       to groups that are associated with hands outside
       our group of interest.
    We use a white bounding box in (b) to highlight
       the appearance of a hand that was fully occluded in (a).
    }
    \label{fig:problem_many_hands}
\end{figure}

\section{Introduction}
We study the problem of developing
   a robust method for detecting student
   hands in collaborative
   learning environment
   \cite{celedon2013interdisciplinary}.
Here, we define a collaborative learning environment
   as a small group of students working together
   in a single table as shown in Fig. \ref{fig:problem_many_hands}.
Our goal is to recognize
  writing and typing
  activities over the
  detected hand regions.
We will then use the
  writing and typing
  activities to assess
  student participation.

For robust detection,
   we require that our hand detection
   results are consistent throughout the video,
   implying that we need to deal with occlusions.
Furthermore, we need to reject
   hands that belong to students that 
   belong to other groups, as opposed
   to the collaborative group
   that is closer to the camera
   (see Fig. \ref{fig:problem_many_hands}).
Since our ultimate goal is to apply
   our methods to about 1,000 hours of
   digital videos, we also require
   that our methods are fast.

We also recognize the dynamic aspects
   of the hand detection problem.
First, it is clear that we need to associate
  hands with different people and
  that there is a need to deal with 
  the fact that hands can disappear from
  view due to occlusion (see Fig. \ref{fig:problem_many_hands}).
Second, we note that the same hands assume
  very different appearances throughout the video
  and that there is a need to associate
  their variations with a single instance.
  
We summarize some earlier research on the same problem
   in the M.Sc. thesis by C.J. Darsey \cite{darsey2018hand}.
In her thesis, the author studied the problem of accurate
   hand segmentation over a limited dataset. 
The dataset consisted of $15$ video clips of a 
    maximum duration of $99$ seconds.
While the methods were successful over a limited video dataset,
   it is important to note that we are dramatically
   extending this prior research to long-term detection
   of hand regions over long video segments.
Thus, unlike \cite{darsey2018hand}, the current paper
   also deals with occlusion, rejecting hands outside
   the group, and associating hand regions with different
   students.
We also have an earlier attempt to detect hands
   using deep learning in \cite{jacoby2018context}.
The current paper dramatically extends this prior
   research that was focused on very short video datasets
   without considering occlusion, appearance issues, and associating hands with different people.
We also note that head detection and person recognition has been studied in
  \cite{shi3}, \cite{shi2}, \cite{Tran2021} and \cite{shi2021}.
Human activity classification over cropped regions was
  studied in  \cite{eilar2016distributed1}, \cite{shi2021b} and \cite{VJ2021}.
 In addtion, we note that speech recognition using speaker geometry is studied in \cite{Luis2021}.

The current paper uses transfer learning from
  deep learning methods to provide initial hand
  detection results.
For this initial step, we tested several
  well-known methods.
We tested Faster R-CNN \cite{ren2016faster}, 
  YOLO  \cite{redmon2016you}, 
  and SSD \cite{liu2016ssd}.
We then decided to adopt Faster R-CNN as our baseline
  model due to the fact that it is more widely supported
  within human activity recognition systems.
We then build our system by post-processing the results
  from Faster R-CNN.
More specifically, we project the results over short
  video segments to address occlusion and then develop
  a clustering approach and small area removal to
  identify the students within the 
  current collaborative group, which are
  not addressed by traditional hand tracking methods (e.g., \cite{HandTracking}).
Our approach yields significant improvements over
  the standard use of Faster R-CNN.
  
The rest of the paper is organized into three additional
    sections.
We summarize the methodology in 
   section \ref{sec:methodology}. 
We then present results in section \ref{sec:results}
   and provide concluding remarks in 
   section \ref{sec:conclusion}.

\section{Methodology}\label{sec:methodology}
We summarize our methodology into two sections.
First, we present a summary of our hand detection method.
Second, we present an optimal data augmentation approach
       to extend our ground truth dataset.

\begin{figure}[!h]
\centering
\includegraphics[width=\textwidth]{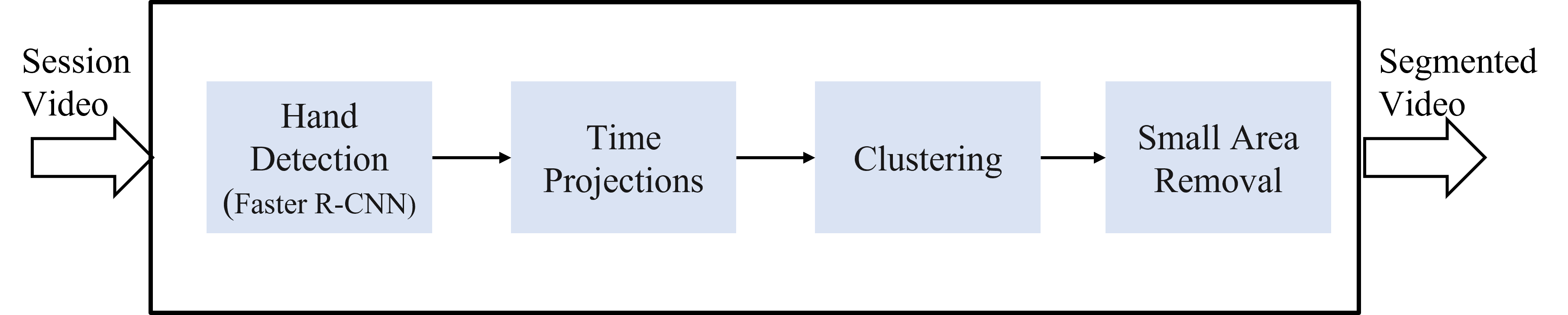}\\
\begin{algorithmic}[0]
\Function{DetectHands}{$w_*, {\tt V}$,  ${\tt a_{\tt th}}$}\\
\COMMENT{{\bf Input}:}\\
\COMMENT{$\quad$ 
   $w_*$ represents a pre-trained 
                  single-frame hand detector.}\\
\COMMENT{$\quad$    
   {\tt V} represents a short Video segment
              of fixed {\tt n} seconds duration.}\\
\COMMENT{$\quad$
    ${\tt a_{\tt th}}$ represents a minimum area requirement.
    }\\
\COMMENT{{\bf Output}:}\\
\COMMENT{$\quad$ 
    {\tt H} contains the detected hand regions for each
            12-second video segment.}
\State {\tt BI} $\gets$ $w_*({\tt V})$
    \COMMENT{detect hands at the rate of one frame per second.}
\State {\tt H} $\gets$ \{\} 
    \COMMENT{initialize {\tt H} to store hand detections.}
\For {each 12-second video segment $i$:}
    \State Project the detected hand regions using:
    \State $\quad {\tt PI_i}$ $\gets$ $\sum_s {{\tt BI_s}}$
    \State Cluster the projected hand regions using:
    \State $\quad {\tt CI_i}$ $\gets$ {\bf Cluster}(${\tt PI_i}$) \State Remove small hand regions of far-away groups:
    \State $\quad {\tt H_i}$ $\gets$ {\bf AreaThreshold}(${\tt CI_i}$, ${\tt a_{\tt th}}$)
    \State {\tt H} $\gets$ {\bf Append}({\tt H}, ${\tt H_i}$)
\EndFor
\State {\bf return} {\tt H}
\EndFunction
\end{algorithmic}
\caption{Proposed hand detection method using time-projections, clustering, and small region removal}\label{fig:method}
\end{figure}

\subsection{Hand detection method}\label{sec:method_our}
\begin{figure}[!t]
\centering
  \subfloat[Hand detections using Faster R-CNN.]{
    \includegraphics[width=0.48\textwidth]{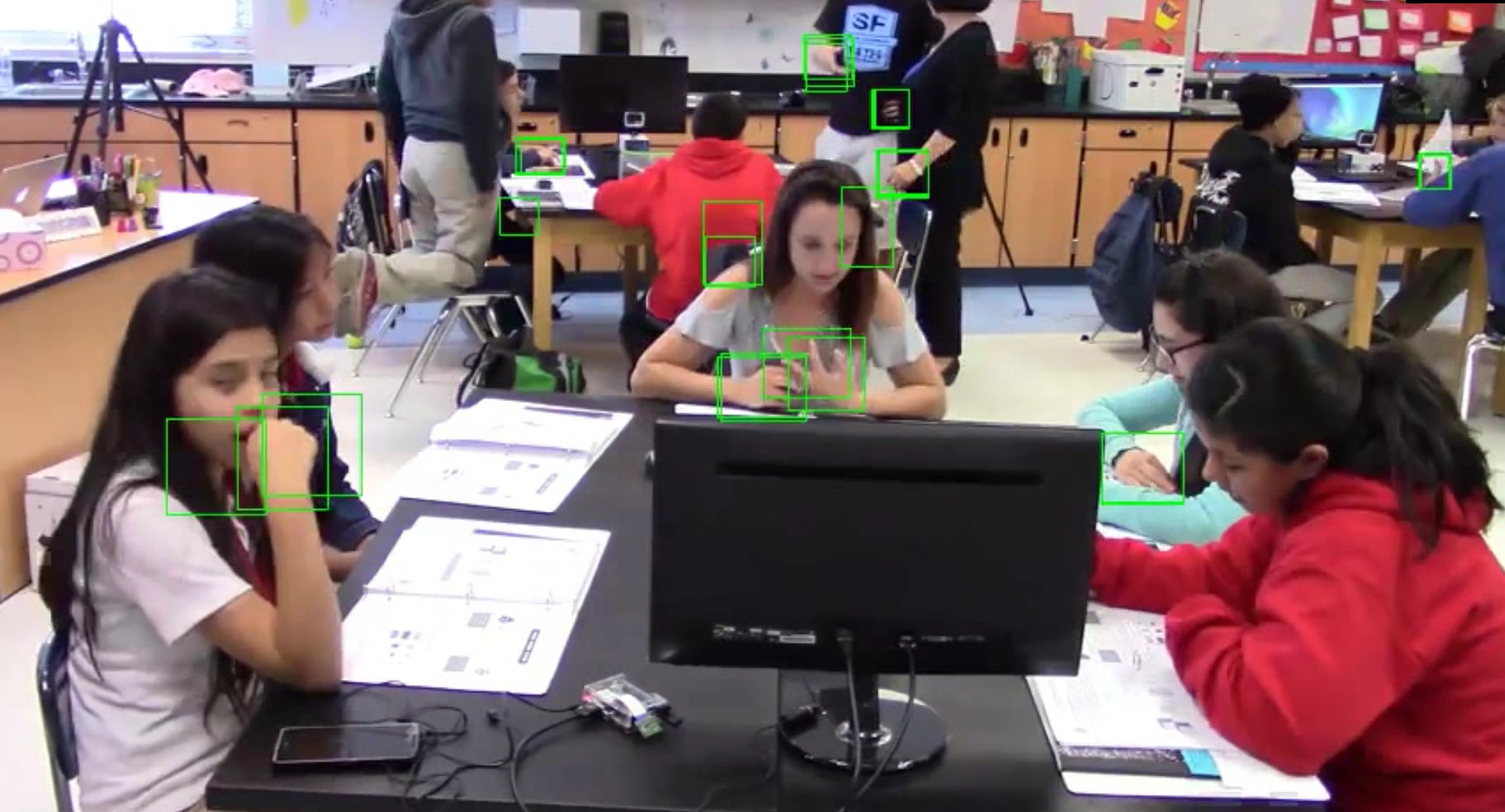}
    \label{subfig:method_faster_rcnn}
  }
  \subfloat[Hand detection projections for $12$ seconds.]{
    \includegraphics[width=0.48\textwidth]{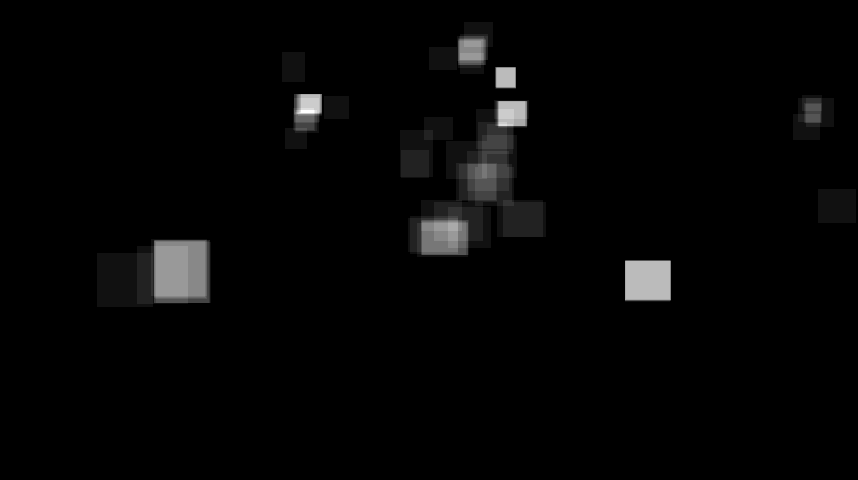}
    \label{subfig:method_projections}
  }

  \subfloat[Binary image showing clusters]{
    \includegraphics[width=0.48\textwidth]{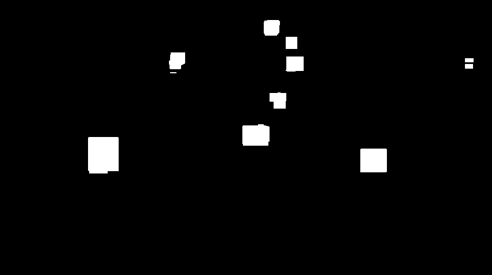}
    \label{subfig:method_cluster_based_seg} 
  }
  \subfloat[Green boxes showing valid clusters after removing small clusters.]{
    \includegraphics[width=0.48\textwidth]{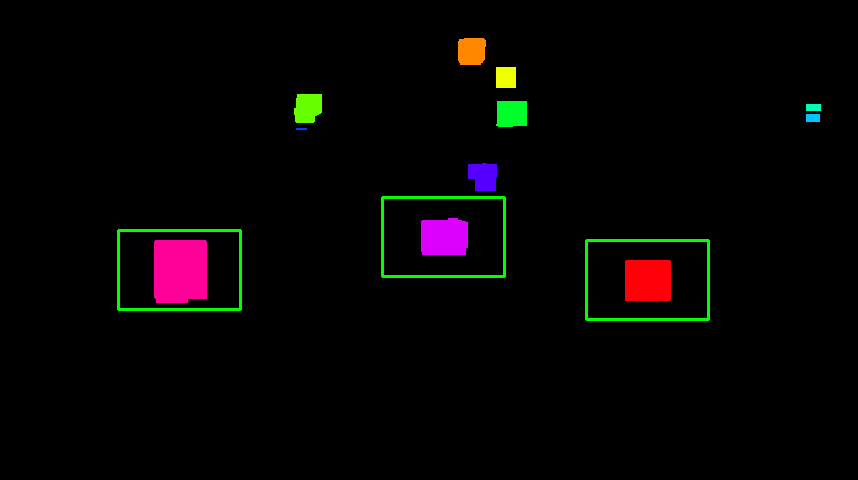}
    \label{subfig:method_small_area_removal}
  }
  
    \subfloat[Hand detections using our method.]{
    \includegraphics[width=0.48\textwidth]{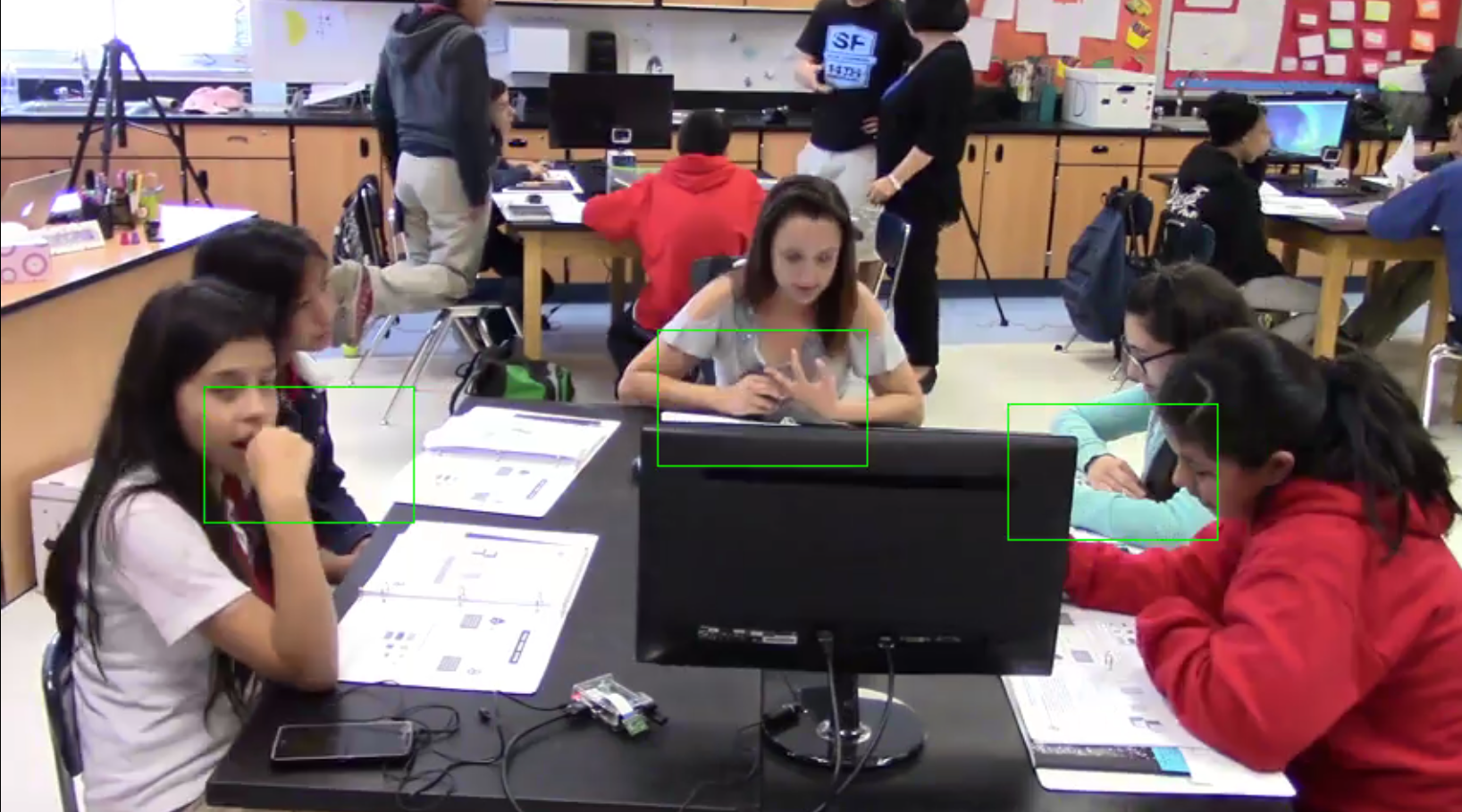}
    \label{subfig:method_final_result}
  }
  \caption{Hand detection images that demonstrate the proposed approach.}
  \label{fig:methodology}
\end{figure}

We present a block diagram and the corresponding pseudo-code of our approach in
   Fig. \ref{fig:method}. 
We begin with a deep-learning method that detects hands at the rate of one frame per second.
The output of the hand detection method is assumed to be 1 over pixel regions that
    represent hand regions, and 0 over other regions.
Then, we take the projection of the detected regions every 12 seconds.
The projected images $\{ {\tt PI_1}, {\tt PI_2}, ..., {\tt PI_{\floor{n/12}}}\}$
    can hold a maximum of 12 that represents hand detection
    over all images, and a minimum of 0 that represents the lack of any hands
    detected over any image.
    
To account for occlusion, appearence, and disappearance,
    we apply a clustering method over the projected image.
Several other standard clustering were investigated during the training process (e.g., Otsu, Li, mean, min, etc \cite{scikit-learn}).
We found that ISODATA \cite{ball1965isodata} performed best.
ISODATA is an iterative method that uses Euclidean distance to determine
    the clusters.

We illustrate the proposed approach in Fig. \ref{fig:methodology}. 
 We show hand detections, obtained after non-maximum supression \cite{non-max-supression}, and clusters using time projections respectively in Figs. \ref{subfig:method_faster_rcnn} and \ref{subfig:method_projections}. Following this,
we were able to reject out of group hand clusters with high confidence based on a
   cluster area constraint \cite{jatla2019image} as shown in Fig. \ref{subfig:method_small_area_removal}.
The final clusters are then shown in Fig. \ref{subfig:method_final_result}.

\subsection{Optimal data augmentation}
\label{sec:method_optimal_data_aug}
For robust detection,
   developed an optimization method
   for augmenting the dataset.
Our goal here is to significantly
   extend the hand dataset
   for different scenarios.

The hand detection dataset was created by extracting frames from $44$ different collaborative learning
sessions. These sessions were selected across $3$ years 
providing a diverse dataset. 
We labeled every hand instance for a total of $4,548$ instances. We partition the dataset into training, validation, and testing
   samples as given in  
   Table \ref{tab:hand_gt_data_split}.
\begin{table}[!h]
  \centering
  \caption{Dataset for training, validation, and testing.
   The training, validation, and testing examples come from different
       video sessions.}
  \label{tab:hand_gt_data_split}
  \begin{tabular}{ l | p{3cm} p{3cm}  p{3cm}}
    & \textbf{\# Sessions} & \textbf{\# Images} & \textbf{\# hand instances} \\
    \hline
    \hline
    \textbf{Training}  & 33 & 305 & 1803 \\
    \textbf{Validation} & 4 & 100 & 714 \\
    \textbf{Testing} & 7 & 313 & 2031 \\
    \hline
    \textbf{Total} & 44 & 718 & 4,548
  \end{tabular}
\end{table}

The ground truth images span multiple video sessions.
For training, we sampled hands from 33 video sessions.
For validation, we sampled hands from another four video sessions.
For testing, we used another set of 7 complete video sessions.
Video sessions were collected over three years.
Video sessions were forty-five
    to one hour and fifteen minutes long.
The training dataset described in
Table \ref{tab:hand_gt_data_split} was carefully selected to have
diversity with $350$ samples. 

We develop a separable optimization approach that 
   starts with determining the maximum range of
   angles for shear, rotation, and pixels to be translated. 
To establish the maximum range of values to consider for 
   shear and rotation, we calculate validation accuracy at multiple angles: 
     $\theta \in \{1^\circ, 2^\circ, 4^\circ, 8^\circ, 16^\circ, 32^\circ\}$.
The maximum range is determined based on the largest angle that
    results in a significant decrease in validation accuracy.
Let [-$\theta^*_r$, $\theta^*_r$], [-$\theta^*_s$, $\theta^*_s$] 
    denote the optimal ranges for rotation and shear, respectively. 
Similarly, we evaluate validation accuracy at multiple horizontal translations:
   $\tau \in \{1, 2, 4, 8, 16, 32, 64, 128,256, 512, 800\}$, 
   and compute the maximum interval: $[-\tau^*,
  \tau^*]$. We summarized augmentation methods, along with their
respective optimal ranges in Table \ref{tab:optimal_aug_ranges}. 

\begin{table}[!t]
   \centering
   \caption{Optimal augmentation parameter value ranges.}
   \label{tab:optimal_aug_ranges}
  \begin{tabular}{p{2.5cm}|p{3cm}}
    Method & Optimal range\\
    \hline
    \hline
    Shear     & $[-3^\circ, 3^\circ]$ \\
    Rotate    & $[-7^\circ, 7^\circ]$ \\
    Translate & $[-20, 20]$ \\
    \hline
  \end{tabular}
\end{table}

In addition to determining the best parameter values for each augmentation method, we also optimize the probability, $\texttt{p}$, for applying data augmentation.
For example, for $\texttt{p}=1$, data augmentation is always applied.
We compute the optimal data augmentation probability $\texttt{p}^*$ as
   described in Fig.  \ref{fig:best_prob_pseudocode}. 

\begin{figure}[!b]
	\begin{algorithmic}[1]
		\For{each $\texttt{p} \in \{0, 0.25, 0.5, 0.75, 1\}$}
		\For{each image in training}
		\State	Apply \textbf{random horizontal flips} with $\texttt{p}$ probability.
		\State	Apply \textbf{random scaling} of \{0.8,1.2\} with $\texttt{p}$ probability.
		\State	Apply \textbf{random shear angle}  sampled from $\{-\theta^*_s,..., \theta^*_s\}$ with $\texttt{p}$ probability.
		\State	Apply \textbf{random rotation angle} sampled from   $\{-\theta^*_r,..., \theta^*_r\}$ 
		\State  $\qquad$ with prbability $\texttt{p}$.
		\State	Apply \textbf{random horizontal translation} with pixels
		\State $\qquad$ uniformly sampled from $\{-\tau^*,..., \tau^*\}$ with $\texttt{p}$ probability.
		\EndFor
		\State \textbf{Train} the model with the augmented data.
		\State \textbf{Record} validation accuracies.
		\EndFor
		\State Select optimal probability ($\texttt{p}^*$) that has the highest validation accuracy 
	\end{algorithmic}
	\caption{Pseudocode for finding the optimal probability for data augmentation.}
	\label{fig:best_prob_pseudocode}
\end{figure}

\section{Results}
\label{sec:results}
We present the results in two sections. 
We first
    present improvement in hand detection by using
    optimal data augmentation method described in
    section \ref{sec:method_optimal_data_aug}. 
We then present the final detetion results that demonstrate that our method
    reduced the number of false positive regions by 
    78.8\% without sacrificing any true positive detections.
    
We used an Intel Xeon 4208 CPU @ 2.10GHz
server, having 128 GB DDR4 RAM and an NVIDIA RTX 5000 GPU for all the experiments. For training Faster
R-CNN, we used the recommended learning rate of $0.001$ for $12$ epochs with a mini-batch size of $2$ images. We can train the model in less than $13$ minutes.

\subsection{Results for optimal data augmentation}
\label{sec:results_opt_data_aug}

Table \ref{tab:optimal_aug_ranges} provides the optimal maximum range angles for
    shear, rotation, and pixels to be translated for hand detection. 
We applied the optimal augmentation values at different probabilities
    as summarized in table \ref{tab:hand_det_results}. 
From this table, it is clear that $0.5$ probability provided the best performance. 
 
\begin{table}[t]
  \centering
  \caption{Hand detection validation and testing average
    precision. From the table, it is clear that $\texttt{p}$ of $0.5$
    gave the best performance.}
  \label{tab:hand_det_results} 
  \begin{tabular}{ p{2cm}  p{1.5cm}| p{1.5cm}  p{1.5cm} p{1.5cm} p{1.5cm} p{1.5cm} }
  \textbf{Data split}&\textbf{Model} & \multicolumn{5}{c}{\textbf{Probability of applying each data augmentaion}}\\
    & & \textbf{0} & \textbf{0.25} & \textbf{0.5} & \textbf{0.75} & \textbf{1.0}\\
    \hline 
    \hline
    \textbf{Val} & Best & 0.77 & 0.86 & \textbf{0.86} & 0.85 & 0.84 \\ 
    & Last & 0.76 & 0.85 & \textbf{0.86} & 0.84 & 0.82 \\
    \textbf{Test} & Best & 0.75 & 0.80 & \textbf{0.80} & 0.79 & 0.78 \\
    & Last & 0.71 & 0.80 & \textbf{0.81} & 0.78 & 0.76 \\
    \hline
  \end{tabular}	
\end{table}

\subsection{Hand detection results}

We summarize our results in table \ref{tab:hand_prop_perf}.
Compared to Faster R-CNN, our approach 
   reduced the number of false positives by $80\%$ while improving IoU ratios from 0.2 to 0.5.
Overall, our hand detector achieved average precision (AP) of $72\%$ at 0.5
    intersection over union (IoU). 
The detection results were improved to
    $81\%$ by using our optimized approach for data augmentation. 
Our method runs at $4.7\times$the real-time.

\begin{figure}[!h]
  \subfloat[Initial hand regions detected using Faster RCNN.]{
    \includegraphics[width=0.49\textwidth]{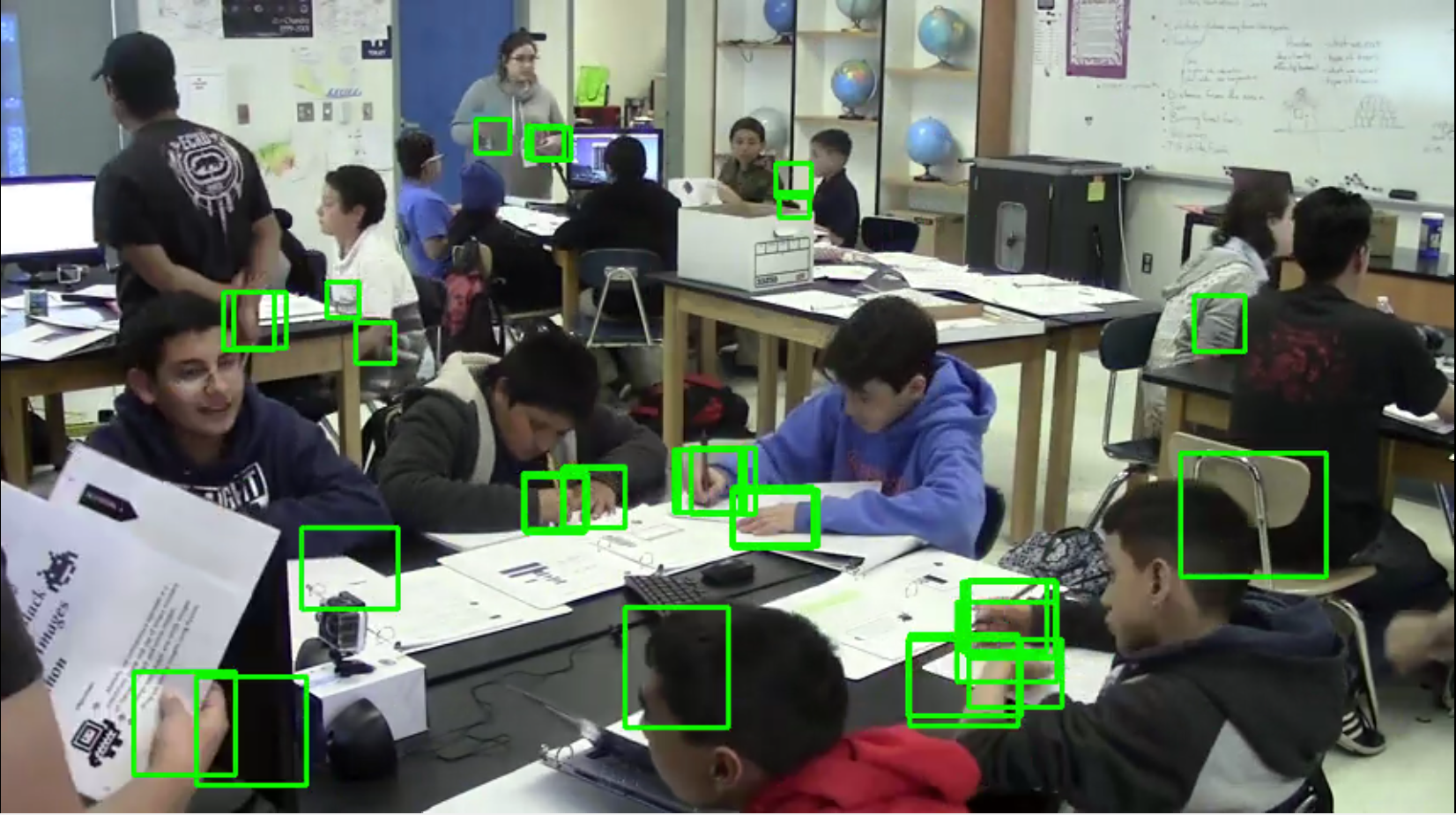}
    \label{subfig:hand_detections_1}
  }\hspace{2mm}
  \subfloat[Ours.]{
    \includegraphics[width=0.49\textwidth]{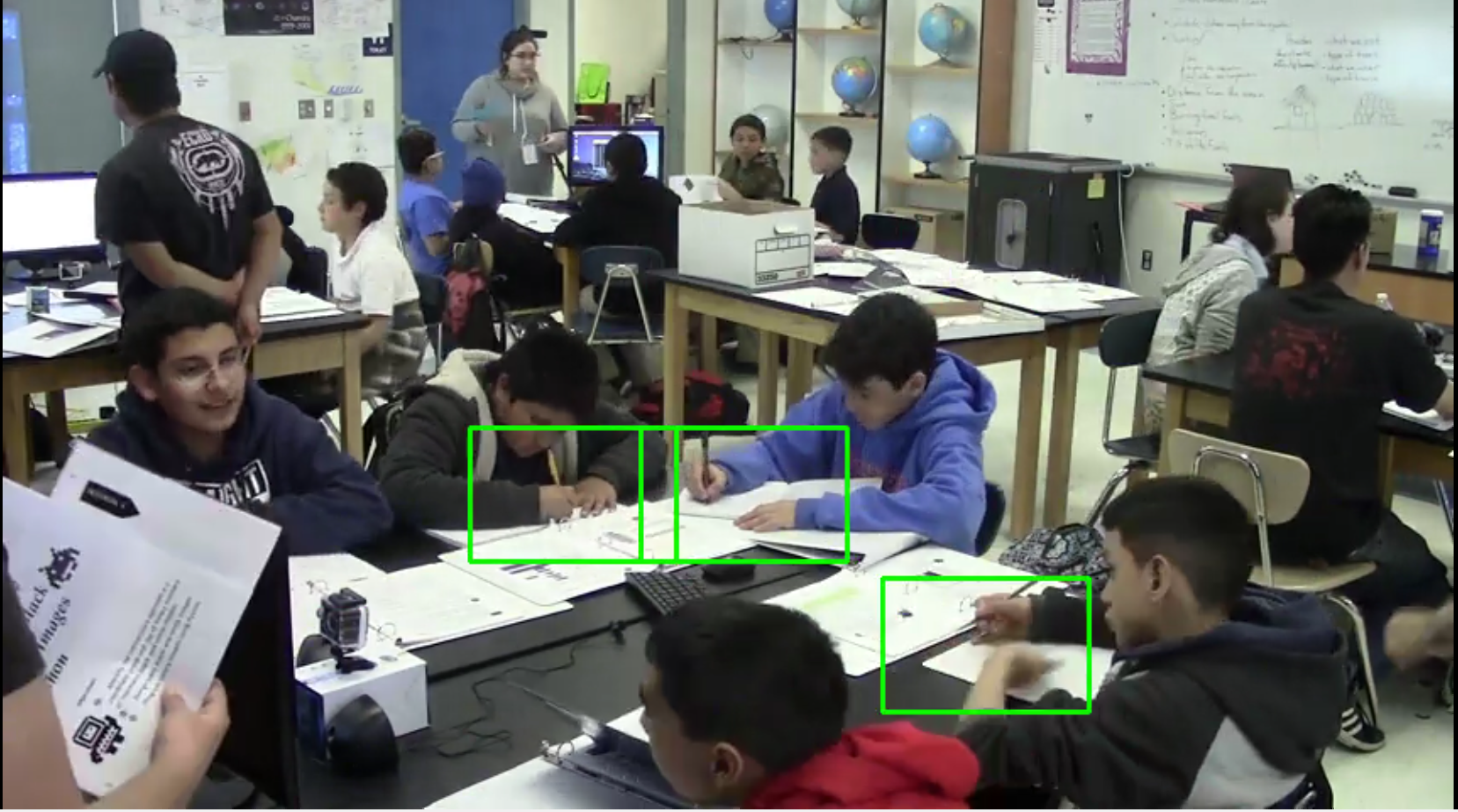}
    \label{subfig:hand_region_prop_1}
  }
  
  \subfloat[Initial hand regions detected using Faster RCNN.]{
    \includegraphics[width=0.49\textwidth]{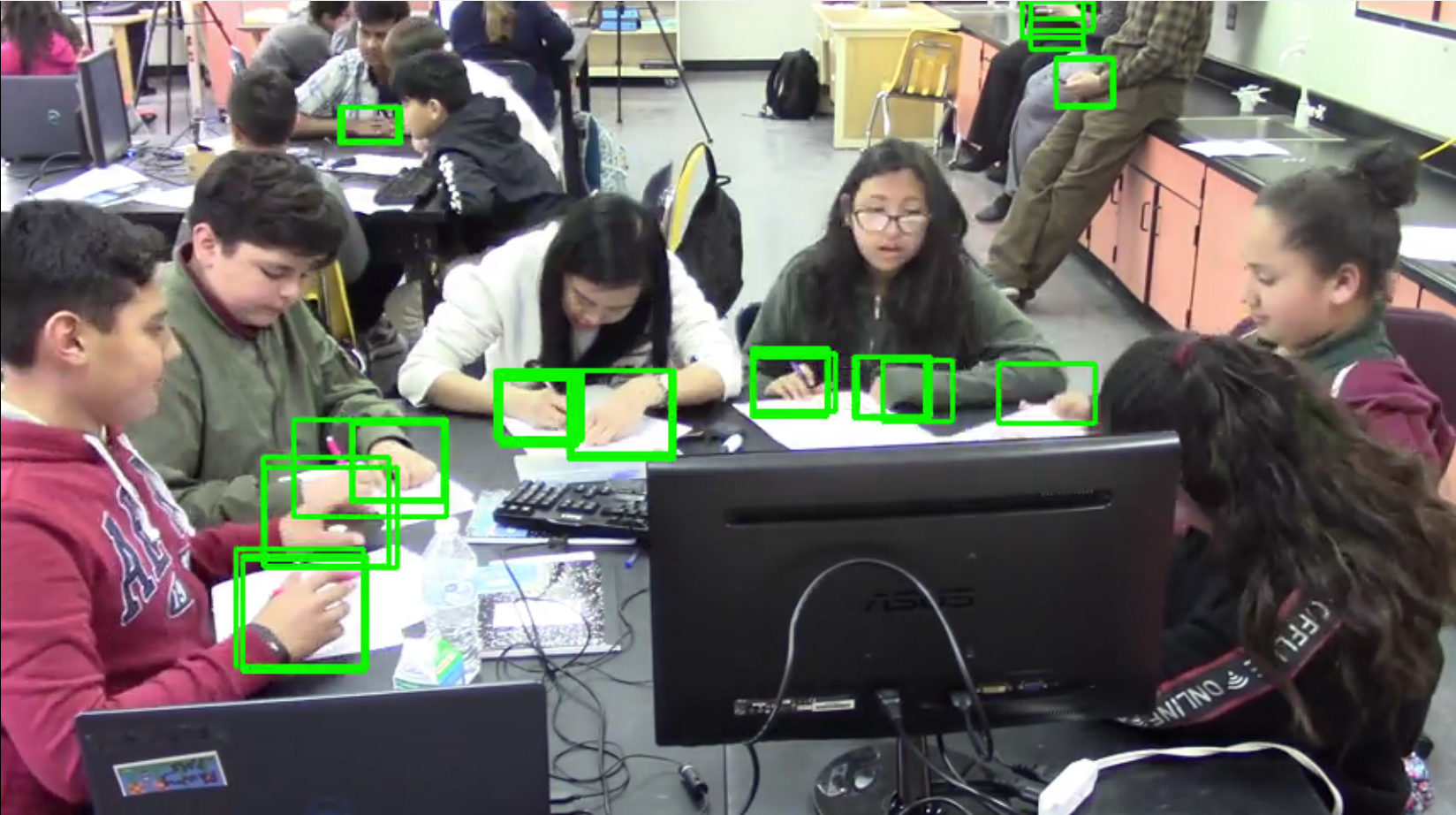}
    \label{subfig:hand_detections_2} 
  }\hspace{2mm}
  \subfloat[Ours.]{
    \includegraphics[width=0.49\textwidth]{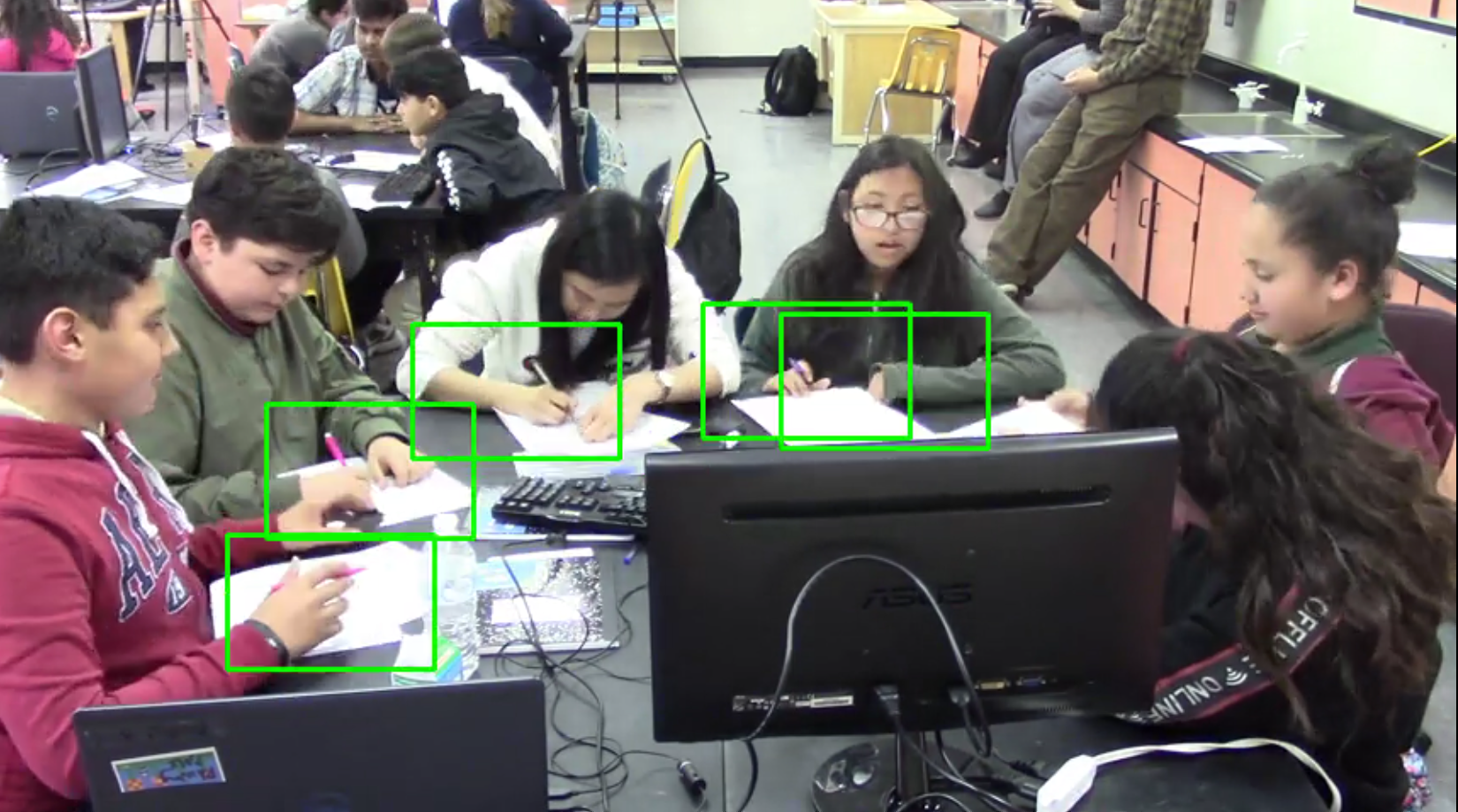}
    \label{subfig:hand_region_prop_2}
  }
  
    \subfloat[Initial hand regions detected using Faster RCNN with significant hand movements.]{
    \includegraphics[width=0.49\textwidth]{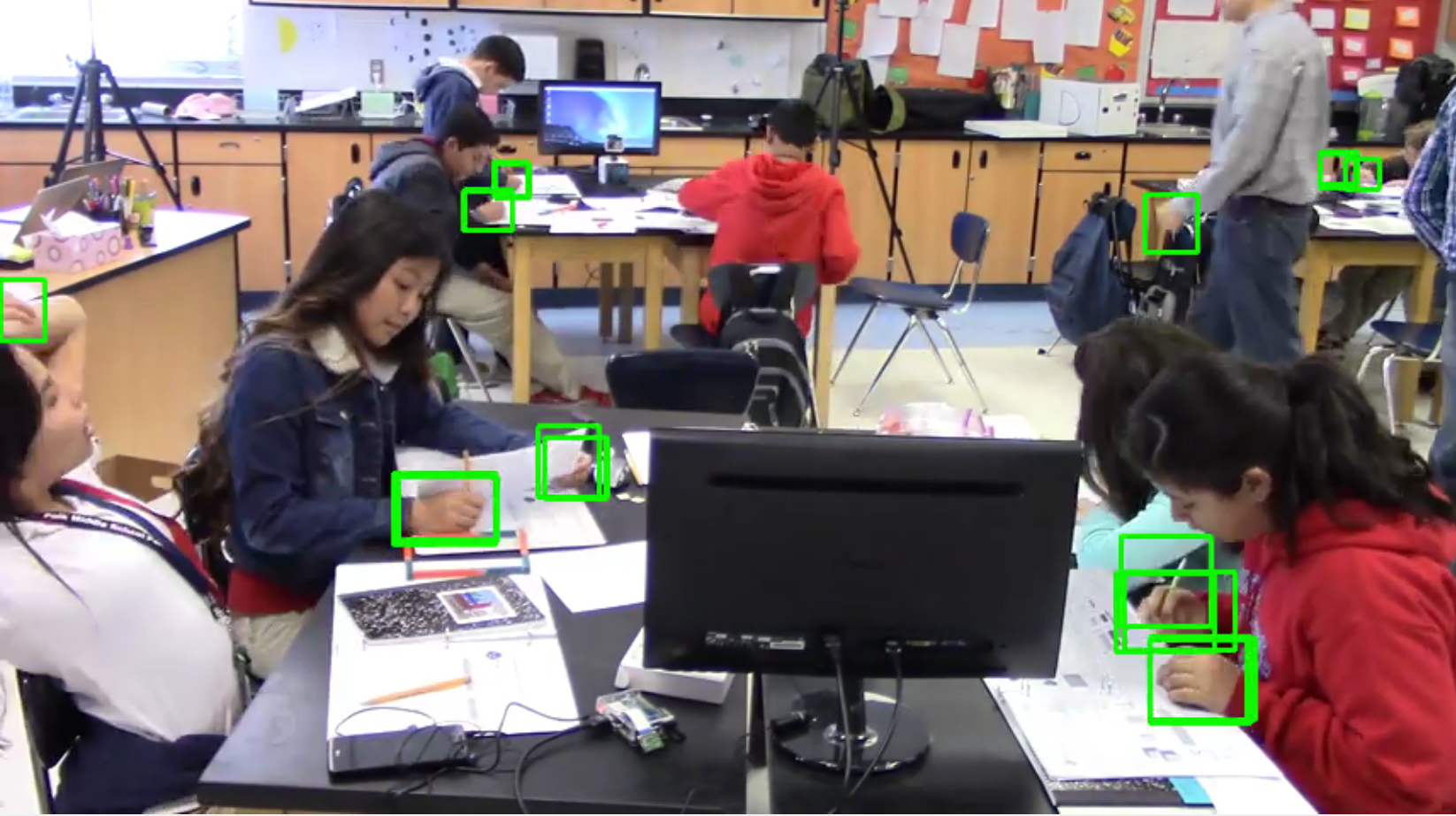}
    \label{subfig:hand_detections_3}
  }\hspace{2mm}
  \subfloat[Ours.]{
    \includegraphics[width=0.49\textwidth]{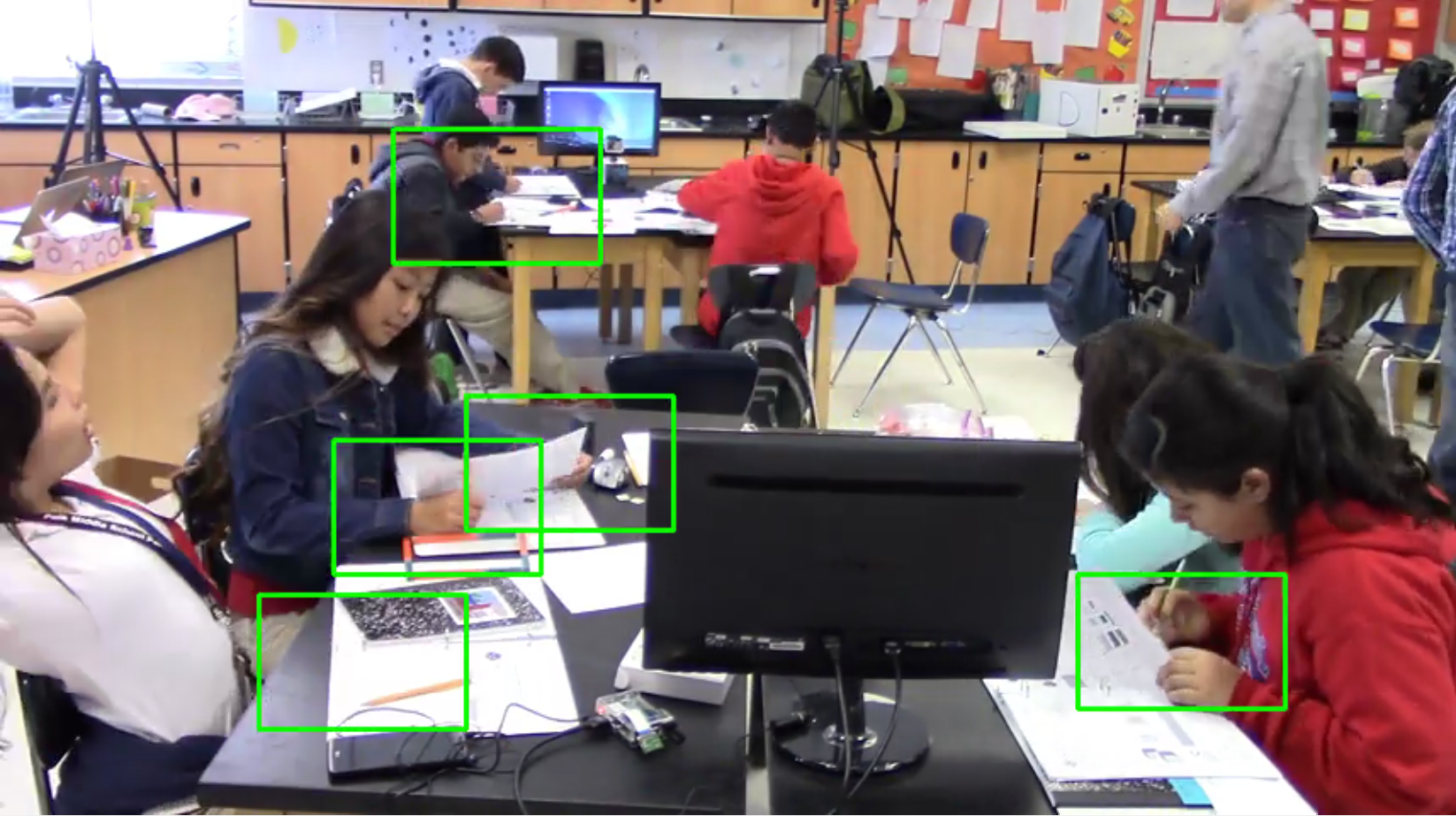}
    \label{subfig:hand_region_prop_3}
  }
  \caption{Comparison between Faster RCNN (left column)
  and our proposed approach (right column).}
  \label{fig:hand_detection}
\end{figure}

We present results against Faster R-CNN in Fig. \ref{fig:hand_detection}.
Overall, we can see that our approach results in a significant
    reduction in the number of detected hand regions.
In some instances, our approach produces
    two overlapping hand regions that are associated with the same student.
    \begin{table}[!t]
  \centering
  \caption{Reduction in number of hand detections for each test session.}
  \label{tab:hand_prop_perf}
  \begin{tabular}{ p{2.75cm} |  p{1.5cm} l p{1.5cm} p{1cm} l}
  \textbf{Session} & \multicolumn{2}{c}{\textbf{\# Hand detections}} & \multicolumn{2}{c}{\textbf{Median IoU}} \\
    & \textbf{Faster RCNN} & \textbf{Ours} & \textbf{Faster RCNN}	& \textbf{Ours} &\textbf{Reduction} \\
    \hline 
    \hline
    C1L1P-C, Mar30 & 55,914 & 9,804  & 0.22 & 0.38 & 82.5\%\\ 
    C1L1P-C, Apr13 & 34,665 & 8,028  & 0.18 & 0.45 & 76.8\%\\
    C1L1P-E, Mar02 & 50,312 & 9,968  & 0.15 & 0.46 & 80.0\%\\
    C2L1P-B, Feb23 & 48,073 & 9,924  & 0.22 & 0.47 & 79.3\%\\
    C2L1P-D, Mar08 & 31,875 & 7,724  & 0.27 & 0.40 & 75.7\%\\
    C3L1P-C, Apr11 & 36,757 & 9,536  & 0.23 & 0.43 &  74.0\%\\
    C3L1P-D, Mar19 & 57,319 & 9,536  & 0.23 & 0.54 &  83.3\%\\
    \hline
  \end{tabular}	
\end{table}

\section{Conclusion}\label{sec:conclusion}
We presented a fast and robust method
   for detecting hands in collaborative learning
   environments.
Our method performed significantly
   better than the standard use of Faster R-CNN.
In future work, the detected proposal regions will be used for the accurate detection of writing and typing activities which can inform educational researchers identify moments of interest in collaborative learning environments.
\clearpage
%
%
\bibliographystyle{splncs04}
\bibliography{SVJ_CAIP_2021}
%




\end{document}